\documentclass{article} 
\usepackage{iclr2015,times}
\usepackage{hyperref}
\usepackage{url}
\usepackage{amsmath, graphicx, subfigure}

\title{Learning linearly separable features for speech recognition using convolutional neural networks}

\author{
Dimitri Palaz \\
Idiap Research Institute, Martigny, Switzerland \\
Ecole Polytechnique F\'ed\'erale de Lausanne (EPFL), Lausanne, Switzerland\\
\texttt{dimitri.palaz@idiap.ch} \\
\And
Mathew Magimai.-Doss \& Ronan Collobert \\
Idiap Research Institute, Martigny, Switzerland \\
\texttt{mathew@idiap.ch, ronan@collobert.com} \\
}

\iclrfinalcopy 

\begin{document}

\maketitle

\begin{abstract}
Automatic speech recognition systems usually rely on spectral-based features, such as MFCC of PLP. These features are extracted based on prior knowledge such as, speech perception or/and speech production. Recently, convolutional neural networks have been shown to be able to estimate phoneme conditional probabilities in a completely data-driven manner, i.e. using directly temporal raw speech signal as input. This system was shown to yield similar or better performance than HMM/ANN based system on phoneme recognition task and on large scale continuous speech recognition task, using less parameters. Motivated by these studies, we investigate the use of simple linear classifier in the CNN-based framework. Thus, the network learns linearly separable features from raw speech. We show that such system yields similar or better performance than MLP based system using cepstral-based features as input. 
\end{abstract}

\section{Introduction}
\label{sec:intro}

State-of-the-art automatic speech recognition (ASR) systems typically divide the task into several sub-tasks, which are
optimized in an independent manner~\citep{bourlard1994connectionist}. In a first step, the data is
transformed into features, usually composed of a dimensionality reduction
phase and an information selection phase, based on the task-specific
knowledge of the phenomena.  These two phases have been carefully
hand-crafted, leading to state-of-the-art features such as mel frequency cepstral coefficients (MFCCs)
or perceptual linear prediction cepstral features (PLPs). In a second step, the likelihood of subword units such as, phonemes is
estimated using generative models or discriminative models. In a final step, dynamic programming techniques
are used to recognize the word sequence given the lexical and syntactical constraints.

Recently, in the hybrid HMM/ANN framework~\citep{bourlard1994connectionist}, there has been growing interests in using ``intermediate'' representations, like short-term spectrum, instead of conventional features, such as cepstral-based features. Representations such as Mel filterbank output or log spectrum have been proposed in the context of deep neural networks~\citep{hinton_deep_2012}. In our recent study~\citep{Palaz_INTERSPEECH_2013}, it was shown that it is possible to estimate phoneme class conditional probabilities by using temporal raw speech signal as input to convolutional neural networks~\citep{lecun_generalization_1989}  (CNNs). This system yielded similar or better results on TIMIT phoneme recognition task with standard hybrid HMM/ANN systems. We also showed that this system is scalable to large vocabulary speech recognition task~\citep{palaz2015Convolutional}. In this case, the CNN-based system was able to outperform the HMM/ANN system with less parameters. 

In this paper, we investigate the features learning capability of the CNN based system with simple classifiers. More specifically, we replace the classification stage of the CNN based system, which was a non-linear multi-layer perceptron, by a linear single layer perceptron. Thus, the features learned by the CNNs are trained to be linearly separable. We evaluate the proposed approach on phoneme recognition task on the TIMIT corpus and on large vocabulary continuous speech recognition on the WSJ corpus. We compare our approach with conventional HMM/ANN system using cepstral-based features. Our studies show that the CNN-based system using a linear classifier yields similar or better performance than the ANN-based approach using MFCC features, with fewer parameters.

The remainder of the paper is organized as follows. Section~\ref{sec-motiv} presents the motivation of this work. Section~\ref{sec-system} presents the architecture of the proposed system. Section~\ref{sec-setup} presents the experimental setup and Section~\ref{sec-results} presents the results. Section~\ref{sec-conclusion} presents the discussion and conclude the paper.

\section{Motivation}
\label{sec-motiv}
 \begin{figure*}[t!]
\begin{center}
\subfigure[][MFCC and PLP extraction pipelines \label{fig:feata}]{\includegraphics[width=1\textwidth]{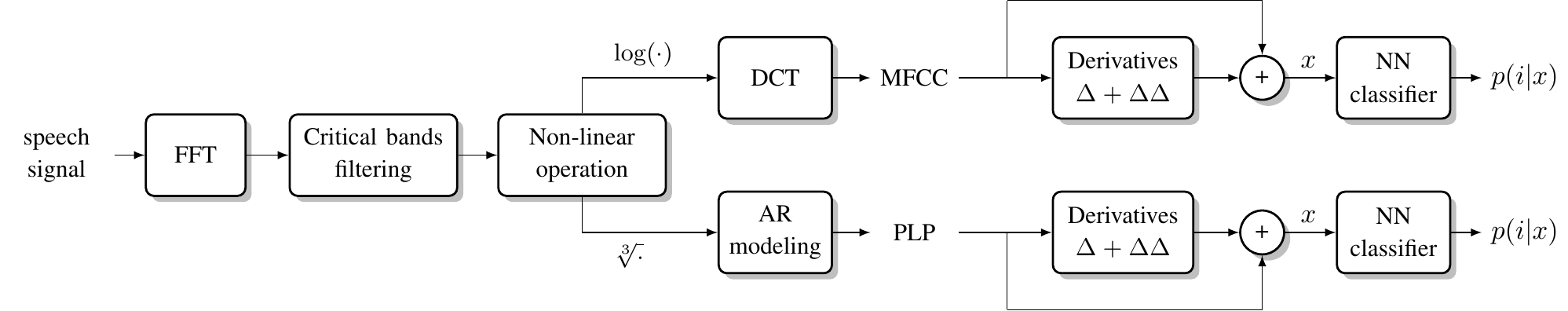}}
\subfigure[][Typical CNN based pipeline using Mel filterbank~\citep{sainath2013deep,Swietojanski:SPL14}\label{fig:featb}]{\includegraphics[width=0.9\textwidth]{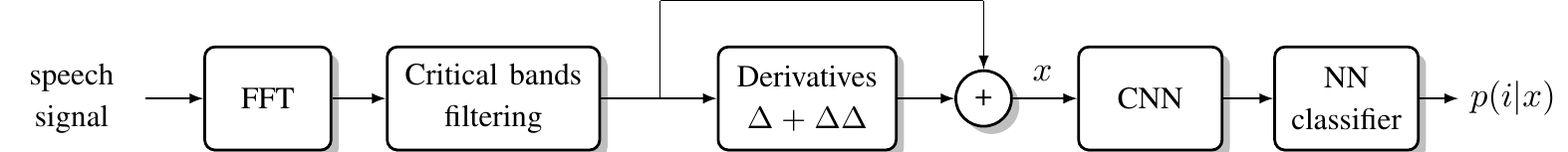}}
\subfigure[][Proposed approach\label{fig:featc}]{\includegraphics[width=0.4\textwidth]{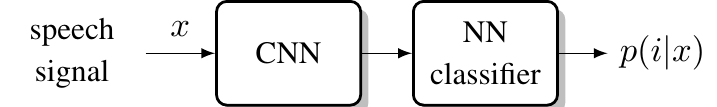}}
\caption{Illustration of several features extraction pipelines. $p(i|x)$ denotes the conditional probabilities for each input frame $x$, for each label $i$.}
\label{fig:feat}
\end{center}
\end{figure*} 
In speech recognition, designing relevant features is not a trivial task, mainly due to the fact that the speech signal is non-stationary and that relevant information is present at different level, namely spectral level and temporal level. Inspired by speech coding studies, feature extraction typically involves modeling the envelop of the short-term spectrum. The two most common features along that line are Mel frequency cepstral coefficient (MFCC)~\citep{davis_comparison_1980} and perceptual linear prediction cepstral coefficient (PLP)~\citep{hermansky1990perceptual}. These features are both based on obtaining a good representation of the short-term power spectrum. They are computed following a series of steps, as presented in Figure~\ref{fig:feata}. The extraction process consists of (1) transforming the temporal data in the frequency domain, (2) filtering the spectrum based on critical bands analysis, which is derived from speech perception knowledge, (3) applying a non-linear operation and (4) applying a transformation to get reduced dimension decorrelated features. This process only models the local spectral level information, on a short time window. To model the temporal variation intrinsic in the speech signal, dynamic features are computed by taking the first and second derivative of the static features on the longer time window, and concatenate them together. These resulting features are then fed to the acoustic modeling part of the speech recognition system, which can be based on Gaussian mixture model (GMM) or artificial neural networks (ANN). In the case of neural networks, the classifier outputs the conditional probabilities $p(i|x)$, with $x$ denoting the input feature and $i$ the class.

In recent years, deep neural network (DNN) based and deep belief network (DBN) based approaches have been proposed~\citep{hinton_fast_2006}, which yield state-of-the-art results in speech recognition using neural networks composed of many hidden layers. In the case of DBN, the networks are initialized in an unsupervised manner. While this original work relied on MFCC features, several approaches have been proposed to use `intermediate'' representations (standing between raw signal and ``classical'' features such as cepstral-based features) as input. In other words, these are approaches that discard several operations in the extraction pipeline of the conventional features  (see Figure~\ref{fig:featb}). For instance, Mel filterbank energies were used as input of convolutional neural networks based \mbox{systems~\citep{abdel-hamid_applying_2012, sainath2013deep,Swietojanski:SPL14}}. Deep neural network based systems using spectrum as input has also been \mbox{proposed~\citep{mohamed_acoustic_2012,lee_unsupervised_2009, sainath_learning_2013}}. Combination of different features has also been \mbox{investigated~\citep{bocchieri2013investigating}}. 

Learning features directly from the raw speech signal using neural networks-based systems has been investigated. In~\cite{jaitly_learning_2011}, the learned features by a DBN are post-processed by adding their temporal derivatives and used as input for another neural network. A recent study investigated acoustic modeling using raw speech as input to a DNN~\cite{tuske2014acoustic}. The study showed that raw speech based system is outperformed by spectral feature based system.
In our recent studies~\citep{Palaz_INTERSPEECH_2013,palaz2015Convolutional}, we showed that it is possible to estimate phoneme class conditional probabilities by using temporal raw speech signal as input to convolutional neural networks (see Figure~\ref{fig:featc}). This system is composed of several filter stages, which performs the features learning step and which are implemented by convolution and max-pooling layers, and of a classification stage, implemented by a multi-layer perceptron. Both stages are trained jointly. On phoneme recognition and on large vocabulary continuous speech recognition task, we showed that the system is able to learn features from the raw speech signal, and yielded performance similar or better than conventional ANN based system that takes cepstral features as input. The proposed system needed less parameters to yield similar performance with conventional systems, suggesting that the learned features seems to be somehow more efficient than cepstral-based features.

Motivated by these studies, the goal of the present paper is to ascertain the capability of the convolutional neural network based system to learn linearly separable features in a data-driven manner. To this aim, we replace the classifier stage of the CNN-based system, which was a non-linear multi-layer perceptron, by a linear single layer perceptron. Our objective is not to show that the proposed approach yields state-of-the-art performance, rather show that learning features  in a data-driven manner together with the classifier leads to flexible features. Using these features as input for a linear classifier yields better performance than SLP-based baseline system and almost reach the performance of MLP-based system.


\section{Convolutional Neural Networks}
\label{sec-system}
This section presents the architecture used in the paper. It is similar to the one presented \mbox{in~\citep{Palaz_INTERSPEECH_2013}}, and is presented here for the sake of clarity.

\subsection{Architecture}
\begin{figure*}[t!]
\begin{center}
\includegraphics[width=0.8\textwidth]{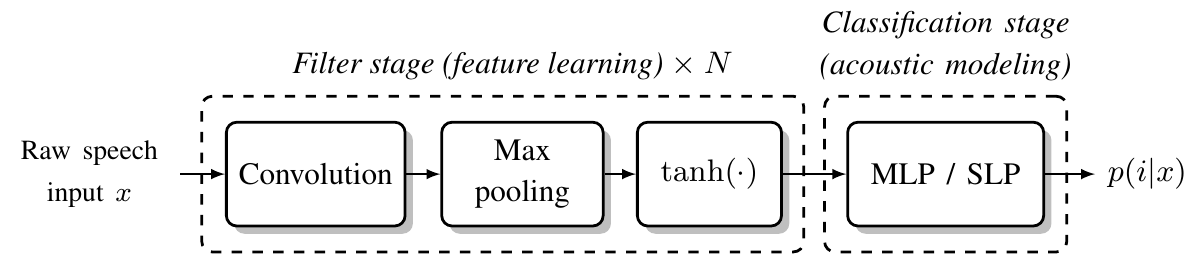}
\caption{ Convolutional neural network based architecture, which estimates the conditional probabilities $p(i|x)$ for each class $i$, for each frame $x$. Several stages of convolution/pooling/tanh might be
considered. The classification stage can be a multi-layer perceptron or a single layer perceptron.}
\label{fig:net}
\end{center}
\end{figure*}

Our network (see Figure~\ref{fig:net}) is given a sequence of raw input signal, split into frames, and outputs a score for each classes, for each frame. The network architecture is
composed of several filter stages, followed by a classification
stage. A filter stage involves a convolutional layer, followed
by a temporal pooling layer and a non-linearity ($\tanh()$).
Processed signal coming out of these stages are fed to a classification stage,
which in our case can be either a multi-layer perceptron (MLP) or a linear single layer perceptron (SLP). It outputs the conditional probabilities $p(i|x)$ for each class $i$, for each frame $x$.

\subsection{Convolutional layer}
While ``classical'' linear layers in standard MLPs accept a fixed-size input vector,
a convolution layer is assumed to be fed with a sequence of $T$ vectors/frames:
$X = \{x^1 \;\; x^2 \;\; \ldots \;\; x^T\}$.  A convolutional layer applies the same linear
transformation over each successive (or interspaced by $dW$ frames) windows of $kW$ frames.
For example, the transformation at frame $t$ is formally written as:
\begin{equation}
M \left( 
\begin{array}{c}  
x^{t-(kW-1)/2} \\ \vdots \\ x^{t+(kW-1)/2}
\end{array} 
\right)\,,
\end{equation}
where $M$ is a $d_{out}\times d_{in}$ matrix of parameters.
In other words, $d_{out}$ filters (rows of the matrix M) are applied
to the input sequence.

\subsection{Max-pooling layer}
These kind of layers perform local temporal $\max$ operations over an input
sequence. 
More formally, the transformation at frame $t$ is written as:
\begin{equation}
\underset{t-(kW-1)/2 \leq s \leq t+(kW-1)/2}{\max} \ x_{s}^d  \quad\quad \forall d
\end{equation}
with $x$ being the input, $kW$ the kernel width and $d$ the dimension.
These layers increase the robustness of the network to minor temporal
distortions in the input.

\subsection{SoftMax layer}
\label{sec-softmax}
The $Softmax$ \citep{bridle_probabilistic_1990} layer interprets network
output scores $f_i(x)$ as conditional probabilities, for each class label
$i$:
\begin{equation}
p(i|x)=\frac{e^{f_i(x)}}{\displaystyle \sum_j e^{f_j(x)}}
\end{equation}

\subsection{Network training}
The network parameters $\theta$ are learned by maximizing the log-likelihood $L$, given by:
\begin{equation}
L(\theta)=\displaystyle \sum_{n=1}^N\log(p(i_n|x_n,\theta)) 
\label{eq:like}
\end{equation}
for each input $x$ and label $i$, over the whole training set (composed of $N$ examples), with respect to the parameters of each layer of the network. Defining the \texttt{logsumexp} operation as: $\operatorname{logsumexp}_i(z_i)=\log(\sum_i e^{z_i})$,
the likelihood can be expressed as:
\begin{equation}
L=\log(p(i|x))=f_i(x) - \underset{j}{\operatorname{logsumexp}}(f_j(x))
\label{eq:likeframe}
\end{equation}
where $f_i(x)$ described the network score of input $x$ and class $i$. The log-likelihood is maximized using the stochastic gradient ascent algorithm \citep{bottou_stochastic_1991}. 

\section{Experimental Setup}
\label{sec-setup}
In this paper, we investigate using the CNN-based approach on a phoneme recognition task and on a large vocabulary continuous speech recognition task. In this section, we present the two tasks, the databases, the baselines and the hyper-parameters of the networks.

\subsection{Tasks}
\subsubsection{Phoneme recognition}

As a first experiment, we propose a phoneme recognition study, where the CNN-based system is used to estimate phoneme class conditional probabilities. The decoder is a standard HMM decoder, with constrained duration of 3 states, and considering all phoneme equally probable.

\subsubsection{Large vocabulary speech recognition}
We evaluate the scalability of the proposed system on a large vocabulary speech recognition task on the WSJ corpus. The CNN-based system is used to compute the posterior probabilities of context-dependent phonemes.
The decoder is an HMM. The scaled likelihoods are estimated by dividing the posterior probability by the prior probability of each class, estimated by counting on the training set. The hyper parameters such as, language scaling factor and the word insertion penalty are determined on the validation set.

\subsection{Databases}
\label{sec-corpus}
For the phoneme recognition task, we use the TIMIT acoustic-phonetic corpus. It consists of 3,696 training utterances (sampled at 16kHz) from 462 speakers, excluding the SA sentences. The cross-validation set consists of 400 utterances from 50 speakers. The core test set was used to report the results. It contains 192 utterances from 24 speakers, excluding the validation set. The 61 hand labeled phonetic symbols are mapped to 39 phonemes with an additional garbage class, as presented in \citep{lee_speaker-independent_1989}.

The the large vocabulary speech recognition task, we use the the SI-284 set of the Wall Street Journal (WSJ) corpus~\citep{woodland_large_1994}. It is formed by combining data from WSJ0 and WSJ1 databases, sampled at 16 kHz. The set contains 36416 sequences, representing around 80 hours of speech. Ten percent of the set was taken as validation set. The Nov'92 set was selected as test set. It contains 330 sequences from 10 speakers. The dictionary was based on the CMU phoneme set, 40 context-independent phonemes. 2776 tied-states were used in the experiment. They were derived by clustering context-dependent phones in HMM/GMM framework using decision tree state tying. The dictionary and the bigram language model provided by the corpus were used. The vocabulary contains 5000 words. 

\subsection{Feature input}
For the CNN-based system, we use raw features as input. They are simply composed of a window of the temporal speech signal (hence, $d_{in}=1$ for the first convolutional layer). The speech samples in the window are normalized to have zero mean and unit variance.

We also performed several baseline experiments, with MFCC as input features. They
were computed (with HTK \citep{htk}) using a 25 ms Hamming window on the
speech signal, with a shift of 10 ms. The signal is represented using
13th-order coefficients along with their first and second derivatives,
computed on a 9 frames context.

\subsection{Baseline systems}

We compare our approach with the standard HMM/ANN system using cepstral features. We train a multi-layer perceptron with one hidden layer, referred to as \emph{MLP}, and a linear single layer perceptron, referred to as \emph{SLP}. The system inputs are MFCC with several frames of preceding and following context. We do not pre-train the network. The MLP baseline performance is consistent with other works~\citep{fosler_crandem_2008}.

\subsection{Networks hyper-parameters}
\label{sec-tuning}

The hyper-parameters of the network are: the input window size $w_{in}$, corresponding to the context taken along with each example, the kernel width $kW_n$, the shift $dW_n$ and the number of filters $d_{n}$ of the $n^{th}$ convolution layer, and the pooling width $kW_{mp}$.
We train the CNN based system with several filter stages (composed of convolution and max-pooling layers). We use between one and five filter stages. In the case of linear classifier, the capacity of the system cannot be tuned directly. It depends on the size of the input of the classifier, which can be adjusted by manually tuning the hyper-parameters of the filter stages. 
The hyper-parameters were tuned by early-stopping on the frame level classification accuracy on the validation set. Ranges which were considered for the grid search are reported in Table \ref{tab:hp}. A fixed learning rate or $10^{-4}$ was used. Each example has a duration of $10$ ms. The experiments were implemented using the \emph{torch7} toolbox \citep{collobert_torch7_2011}. 

On the TIMIT corpus, using 2 filter stages, the best performance was found with: $310$ ms of context, $30$ samples width for the first convolution, $7$ frames kernel width for the second convolution,  $80$ and $60$ filters and $3$ pooling width. Using 3 filter stages, the best performance was found with: $310$ ms of context, $30$ samples width for the first convolution, $7$ and $7$ frames kernel width for the other convolutions,  $80$, $60$ and $60$ filters and $3$ pooling width. Using 4 filter stages, the best performance was found with: $310$ ms of context, $30$ samples width for the first convolution, $7$, $7$ and $7$ frames kernel width for the other convolutions,  $80$, $60$, $60$ and $60$ filters and $3$ pooling width. 
We also set the hyper-parameters to have a fixed classifier input. They are presented in Table~\ref{tab:fixed}. For the baselines, the {\it MLP} uses $500$ nodes for the hidden layer and $9$ frames as context. The {\it SLP} based system uses $9$ frames as context. 

On the WSJ corpus, using 1 filter stage, the best performance was found with: $210$ ms of context, $30$ samples width for the first convolution,  $80$ filters and $50$ pooling width. Using 2 filter stages, the best performance was found with: $310$ ms of context, $30$ samples width for the first convolution, $7$ frames kernel width for the other convolutions,  $80$ and $40$ filters and $7$ pooling width. Using 3 filter stages, the best performance was found with: $310$ ms of context, $30$ samples width for the first convolution, $7$ and $7$ frames kernel width for the other convolutions,  $80$, $60$ and $60$ filters and $3$ pooling width. We also ran experiments using hyper-parameters outside the ranges considered previously using 4 filter stages. This experiment has the following hyper-parameters: $310$ ms of context, $30$ samples width for the first convolution, $25$, $25$ and $25$ frames kernel width for the other convolutions,  $80$, $60$ and $39$ filters and $2$ pooling width. For the baselines, the {\it MLP} uses $1000$ nodes for the hidden layer and $9$ frames as context. The {\it SLP} based system uses $9$ frames as context.

\begin{table}[htb]
\caption{Network hyper-parameters ranges considered for tuning on the validation set.}
\label{tab:hp}
\begin{center}
\begin{tabular}{lll}
 \multicolumn{1}{c}{\bf Hyper-parameter}  &\multicolumn{1}{c}{\bf Units} & \multicolumn{1}{c}{\bf Range} \\ \hline \\
Input window size ($w_{in}$) & ms & 100-700 \\
Kernel width of the first conv. ($kW_1$)& samples & 10-90\\
Kernel width of the $n^{th}$ conv. ($kW_n$) & frames & 1-11\\
Number of filters per kernel ($d_{out}$) & filters &20-100 \\
Max-pooling kernel width ($kW_{mp}$) & frames & 2-6 \\
\end{tabular}
\end{center}
\end{table}

\begin{table}[htb]
\caption{Network hyper-parameters for a fixed output size}
\label{tab:fixed}
\begin{center}
\begin{tabular}{ccccccccccc}
 \multicolumn{1}{c}{\bf \# conv. layer}  &\multicolumn{1}{c}{\bf $w_{in}$} &\multicolumn{1}{c}{\bf $kW_1$} &\multicolumn{1}{c}{\bf $kW_2$} &\multicolumn{1}{c}{\bf $kW_3$} &\multicolumn{1}{c}{\bf $kW_4$} &\multicolumn{1}{c}{\bf $kW_5$} &\multicolumn{1}{c}{\bf $d_{n}$} & \multicolumn{1}{c}{\bf $kW_{mp}$} & \multicolumn{1}{c}{\bf \# output} \\ \hline 
1 & 310 & 3 & na & na & na & na & 39 & 50 & 351 \\
2 & 310 & 3 & 7 & na & na & na & 39 & 7 & 351 \\
3 & 430 & 3 & 5 & 5 & na & na & 39 & 4 & 351 \\
4 & 510 & 3 & 5 & 3 & 3 & na & 39 & 3 & 351 \\
5 & 310 & 3 & 5 & 7 & 7 & 7 & 39 & 2 & 351 \\
\end{tabular}
\end{center}
\end{table}

\section{Results}
\label{sec-results}

The results for the phoneme recognition task on the TIMIT corpus are presented in Table~\ref{tab:res1}. The performance is expressed in terms of phone error rate (PER). The number of parameters in the classifier and in the filter stages are also presented. Using a linear classifier, the proposed CNN-based system outperforms the MLP based baseline with three or more filter stages. It can be observed that the performance of the CNN-based system improves with increase in number of convolution layers and almost approaches the case where a MLP (with 60 more parameters) is used in the classification stages. Furthermore, it can be observed that the complexity of the classification stage decreases drastically with the increase in the number of convolution layers.
The results for the proposed system with a fixed output size is presented in Table~\ref{tab:res2}, along with the baseline performance and the number of the parameters in the classifier and filter stages. The proposed CNN based system outperforms the SLP based baseline with the same number of parameters in the classifier. Fixing the output size seems to degrade the performance compared to Table~\ref{tab:res1}. This indicate that it is better to treat the feature size also as a hyper-parameter and learn it on the data.

\begin{table}[h!]
\caption{Results on the TIMIT core testset}
\label{tab:res1}
\begin{center}
\begin{tabular}{ccccccc}
 & \multicolumn{1}{c}{\bf \# conv.}& \multicolumn{1}{c}{\bf \# conv.}&  & \multicolumn{1}{c}{\bf \# classifier}&  \\
  \multicolumn{1}{c}{\bf Features}  & \multicolumn{1}{c}{\bf layers}& \multicolumn{1}{c}{\bf param.}& \multicolumn{1}{c}{\bf Classifier} & \multicolumn{1}{c}{\bf param.}& \multicolumn{1}{c}{\bf PER} \\ \hline \\
MFCC & na & na & MLP & 200k & 33.3 \% \\
RAW & 3 & 61k & MLP & 470k & 29.6 \% \\
MFCC & na & na & SLP & 14k & 51.5 \% \\ \hline
RAW & 2 & 36k & SLP & 124k & 38.0  \% \\
RAW & 3 & 61k & SLP & 36k & 31.5 \% \\ 
RAW & 4 & 85k & SLP & 7k & 30.2 \% \\\hline
\end{tabular}
\end{center}
\end{table} 

\begin{table}[h!]
\caption{Results for a fixed output on the TIMIT core testset}
\label{tab:res2}
\begin{center}
\begin{tabular}{ccccccc}
 & \multicolumn{1}{c}{\bf \# conv.}& \multicolumn{1}{c}{\bf \# conv.}&  & \multicolumn{1}{c}{\bf \# classifier}&  \\
  \multicolumn{1}{c}{\bf Features}  & \multicolumn{1}{c}{\bf layers}& \multicolumn{1}{c}{\bf param.}& \multicolumn{1}{c}{\bf Classifier} & \multicolumn{1}{c}{\bf param.}& \multicolumn{1}{c}{\bf PER} \\ \hline \\
 MFCC & na & na & SLP & 14k & 51.5 \% \\\hline 
RAW & 1 & 1.2k & SLP & 14k & 49.3\% \\
RAW & 2 & 24k & SLP & 14k & 38.0  \% \\
RAW & 3 & 152k & SLP & 14k & 33.4 \% \\ 
RAW & 4 & 270k & SLP & 14k & 34.6 \% \\ 
RAW & 5 & 520k & SLP & 14k & 33.1 \% \\\hline 
\end{tabular}
\end{center}
\end{table} 
 
The results for the large vocabulary continuous speech recognition task on the WSJ corpus are presented in Table~\ref{tab:res3}. The performance is expressed in term of word error rate (WER). We observe a similar trend to the TIMIT results, i.e. with the increase in number of convolution layers the performance of the system improves. More specifically, it can be observed that with only two convolution layers the proposed system is able to achieve performance comparable to SLP-based system with MFCC as input. With three convolution layers the proposed system is approaching the MLP-based systems. With four convolution layers, the system is able to yield similar performance with the MLP baseline using MFCC as input.  

Overall, it can be observed that the CNN-based approach can lead to systems with simple classifiers, i.e. with a small number of parameters, thus shifting the system capacity to the feature learning stage of the system. On the phoneme recognition study (see Table~\ref{tab:res1}), the proposed approach even leads to a system where most parameters lie in the feature learning stage rather than in the classification stage. This system yields performance similar to or better than baselines system. On the continuous speech recognition study, it can be observed that the four convolution layers experiment has five times less parameters in the classifier than the three layers experiment and still yields better performance. This four layers experiement is also able to yield similar performance to the MLP-based baseline with two times less parameters.


\begin{table}[h!]
\caption{Results on the Nov'92 testset of the WSJ corpus.}
\label{tab:res3}
\begin{center}
\begin{tabular}{ccccccc}
 & \multicolumn{1}{c}{\bf \# conv.}& \multicolumn{1}{c}{\bf \# conv.}&  & \multicolumn{1}{c}{\bf \# classifier}&  \\
  \multicolumn{1}{c}{\bf Features}  & \multicolumn{1}{c}{\bf layers}& \multicolumn{1}{c}{\bf param.}& \multicolumn{1}{c}{\bf Classifier} & \multicolumn{1}{c}{\bf param.}& \multicolumn{1}{c}{\bf WER} \\ \hline \\
MFCC & na & na & MLP & 3M & 7.0 \% \\
RAW & 3 & 55k & MLP & 3M & 6.7 \% \\ 
MFCC &  na & na & SLP & 1M & 10.9 \% \\\hline
RAW & 1 & 5k & SLP & 1.3M & 15.5 \% \\
RAW & 2 & 27k & SLP & 1M & 10.5 \% \\
RAW & 3 & 64k & SLP & 2.4M & 7.6 \% \\
RAW & 4 & 180k & SLP & 410k & 6.9 \% \\\hline
\end{tabular}
\end{center}
\end{table}

\section{Discussion and conclusion}
\label{sec-conclusion}

Traditionally in speech recognition systems, feature extraction and acoustic modeling (classifier training) are dealt in two separate steps, where feature extraction is knowledge-driven, and classifier training in data-driven. In the CNN-based approach with raw speech signal as input, both feature extraction and classifier training is data-driven. Such an approach allows the features to be flexible as they are learned along with the classifier. It also allows to shift the system capacity from the classifier stage to the feature extraction stage of the system. Our studies indicate that these empirically learned features can be linearly separable and could yield systems that perform similar to or better than standard spectral-based systems. This can have potential implication for low resource speech recognition. This is part of our future investigation.


\subsubsection*{Acknowledgments}
This work was supported by the HASLER foundation (www.haslerstiftung.ch) through the grant ``Universal Spoken Term Detection with Deep Learning'' (DeepSTD).  The authors also thank their colleague Ramya Rasipuram for providing the HMM setup for WSJ.
\bibliographystyle{iclr2015}
\bibliography{biblio.bib}

\end{document}